# Explainable machine learning for predicting shellfish toxicity in the Adriatic Sea using long-term monitoring data of HABs

Martin Marzidovšek[1,4,*], Janja Francé[2], Vid Podpečan[1], Stanka Vadnjal[3], Jožica Dolenc[3], Patricija Mozetič[2]

1 Jožef Stefan Institute, Knowledge Technologies E8 (Slovenia)
2 National Institute of Biology, Marine Biology Station Piran (Slovenia)
3 University of Ljubljana, Veterinary Faculty, National Veterinary Institute (Slovenia)
4 Jožef Stefan International Postgraduate School (Slovenia)

ABSTRACT

In this study, explainable machine learning techniques are applied to predict the toxicity of mussels in the Gulf of Trieste (Adriatic Sea) caused by harmful algal blooms. By analysing a newly created 28-year dataset containing records of toxic phytoplankton in mussel farming areas and toxin concentrations in mussels (*Mytilus galloprovincialis*), we train and evaluate the performance of ML models to accurately predict diarrhetic shellfish poisoning (DSP) events. The random forest model provided the best prediction of positive toxicity results based on the F1 score. Explainability methods such as permutation importance and SHAP identified key species (*Dinophysis fortii* and *D. caudata*) and environmental factors (salinity, river discharge and precipitation) as the best predictors of DSP outbreaks. These findings are important for improving early warning systems and supporting sustainable aquaculture practices.

*Keywords:* harmful algal blooms, DSP toxins, machine learning, explainable artificial intelligence, aquaculture, marine ecology, Adriatic Sea.

## 1. Introduction

Over the last 30 years, shellfish aquaculture has steadily and accounted for 25% of global marine and coastal aquaculture production in 2020, which is crucial for food security (FAO-IOC-IAEA 2023). However, for certain species such as mussels, production in the European Union has declined due to challenges such as low profitability, limited progress in farming and environmental threats such as harmful algal blooms (HABs), diseases, predators, unfavourable weather conditions and pollution (Avdelas et al. 2021). These global challenges, exacerbated by climate change, require common practices and guidelines for their detection and appropriate management, as in the case of HABs.

---

[*] Corresponding author at: Jozef Stefan Institute, Jamova 39, Ljubljana, 1000 Slovenia.
E-mail address: martin.marzidovsek@ijs.si (M. Marzidovšek).

The term HABs refers to either non-toxic microalgae that reach high biomass and cause water discoloration, anoxia and mucilage formation that negatively impact the environment and human activities, or toxic species that threaten the safety of seafood and marine life (reviewed in Sagarminaga et al. 2023). The toxin-producing species that can cause food poisoning with neurological or gastrointestinal symptoms in humans are responsible for about 48% of the documented global HAB events recorded in the HAEDAT database[*]. In these events, toxins accumulate in shellfish and fish, severely impacting livelihoods and food security (Hallegraeff et al. 2021). To address these issues, countries around the world have introduced systems for HAB monitoring and management.

In the Mediterranean region, around three quarters of recorded toxic events involve diarrhetic shellfish poisoning (DSP) caused by dinoflagellates like *Dinophysis, Phalacroma* and *Prorocentrum* (Zingone et al. 2021). Paralytic shellfish poisoning (PSP), which is attributed to dinoflagellate species of the genus *Alexandrium* and *Gymnodinium catenatum*, and amnesic shellfish poisoning (ASP), which is attributed to several toxin-producing diatom species of the genus *Pseudo-nitzschia*, make up the rest of the toxic events (Zingone et al. 2021). The Adriatic Sea has a similar prevalence of DSP events. In Adriatic shellfish, DSP toxins of the okadaic acid group together with other lipophilic toxins such as yessotoxins and pectenotoxins, are most frequently detected and exceed legal limits (Accoroni et al. 2024; Nincevic Gladan et al. 2011), while ASP and PSP toxins represent only a low risk for the time being (Ciminiello et al. 2005; Ujević et al. 2012).

In our area of interest, the Gulf of Trieste (northern Adriatic), farmed Mediterranean mussels (*Mytilus galloprovincialis*) are the main source of potential human poisoning. The monitoring programme for toxins in mussels and toxic phytoplankton in seawater has been ongoing since 1994. Patterns suggest a higher risk of exceeding regulatory limits from September to November consistent with the presence of DSP-producing dinoflagellates (Henigman et al. 2024). Despite robust monitoring systems aligned with EU regulations, complete protection against contaminated seafood cannot be guaranteed. Sales are only suspended once toxicity has been confirmed, while in certain cases even changes in toxic plankton can lead to precautionary closures until toxicity results are available, leading to periods of uncertainty.

Furthermore, the usual patterns of toxic species and toxin occurrence may also be disrupted by the unpredictable temporal variability of HABs, as observed in the Adriatic Sea and possibly related to extreme weather events such as floods and droughts (Zingone et al. 2021). Incorrect or untimely measures pose a risk to human health and may cause undue economic damage to shellfish farmers. Given the negative economic impact, solutions are urgently needed to help the shellfish industry anticipate and adapt to HAB events that may lead to contamination and to assist authorities in managing and mitigating risks such as the closure of shellfish harvests.

A recent comprehensive overview of early warning systems (EWS) for HABs (FAO-IOC-IAEA, 2023), pointed out that there is no single approach. The choice of observational technologies depends on the target organisms, types of HABs, and regions with specific environmental conditions as well as data resolution. For toxic pelagic HABs, where the causative organisms are present in low abundance—such as the case of *Dinophysis* species in the Adriatic Sea—various statistical and rule-based modelling, along with more complex machine learning and deep learning algorithms, demonstrate promising results in predicting toxicity in shellfish (e.g. (Grasso et al. 2019)).

---

[*] https://haedat.iode.org/

Access to vast amounts of physico-chemical and biological data from multiple sources due to technological advances in meteorology and oceanography has led to the increasing use of machine learning (ML) techniques to predict HABs (e.g. Guallar et al. 2016; Derot, Yajima, and Jacquet 2020; Y. Park et al. 2021). ML can handle large and heterogeneous datasets and are powerful enough to model highly dynamic and nonlinear natural systems, making them suitable for accurately reproducing phytoplankton dynamics (Shimoda and Arhonditsis 2016), even when the data are noisy and the underlying relationships are not fully understood (Muttil and Chau 2006). However, a recent review of ML forecasting tools (Cruz et al. 2021) found that, in contrast to forecasting HAB species occurrence, very little has been done to foresee the toxicity of mussels (e.g. (e.g. (Bouquet et al. 2022; Grasso et al. 2019). This more difficult task is addressed in our study.

One limitation of ML techniques is that they often do not provide insights into the causal mechanisms of HABs (Recknagel, Orr, and Cao 2014). Simpler ML algorithms such as decision trees (DT) and linear regression are easier to interpret but cannot provide the required predictive power. Cruz et al. (2021) found that model complexity has increased in recent years, at the expense of explainability. However, when modelling scenarios that involve risks to human health and ecosystem disruption, as is the case with HABs, it is crucial not only to assess the reliability of predictions, but also to understand the drivers behind the model's decisions. For these reasons, this study focuses on the training of interpretable ML models and the application of explainability methods that can provide insights into the behaviour of the model. It thus fits into the paradigm of explainable artificial intelligence (XAI), which has recently gained prominence as it promises to mitigate the drawbacks of so-called "black-box" models.

Our study represents one of the first attempts to test the performance of ML models for the short-term prediction of DSP toxicity of mussels in the Mediterranean Sea, where DSP events are the major concern for seafood safety. A nearly 30-year dataset of phytoplankton species and toxin concentrations in bivalves was created along with key environmental data to train ML models, and both the dataset and the full code are openly available to stimulate further research. The obtained ML models were also evaluated and interpreted based on real, long-term monitoring data.

The objectives of the study were: (1) to develop data preprocessing pipelines suitable for the specific requirements of the dataset; (2) to train selected ML models for direct prediction of DSP toxicity events and evaluate their potential for use in EWS; (3) to explain the obtained ML models and their predictions improve trustworthiness for use.

## 2. Material and Methods

### 2.1. Site description

The Slovenian Sea is situated in the southeastern region of the Gulf of Trieste (GoT), marking the shallowest (average depth around 20 m) and northernmost part of the Adriatic Sea. Local meteorological conditions, nutrient inflows from rivers—primarily the Soča River, prevailing currents, and the exchange of water masses with the northern Adriatic collectively shape the physical, chemical, and biological characteristics of the sea. This dynamic interplay results in significant fluctuations in the measured parameters (Malacic and Petelin 2001), which have been overlaid by the effects of climate change in recent decades. The increasing warming of seawater, alternating droughts and floods, combined with human activities in the watershed, have led to an imbalance of nutrients in coastal waters and a general decline in phytoplankton biomass not only in GoT but throughout the northern Adriatic (Brush et al. 2020).

Moreover, the Slovenian coastal sea faces substantial anthropogenic pressures. The area is highly urbanised, experiencing intensive land-based and nautical tourism, hosting an international cargo port, and witnessing an expanding aquaculture industry. In particular, mussel farming is practised in three protected bays (12-18 m deep) on the Slovenian coast - Debeli rtič, Strunjan and Seča (Figure 1) with annual mussel production up to 700 tonnes[†]. Surveillance of the seafood safety began in the late 1980s on an irregular basis, while the regular national monitoring programme for toxic phytoplankton and toxicity in bivalve molluscs was introduced in 1994.

---

[†] Statistical Office of the Republic of Slovenia: https://www.stat.si/statweb/en

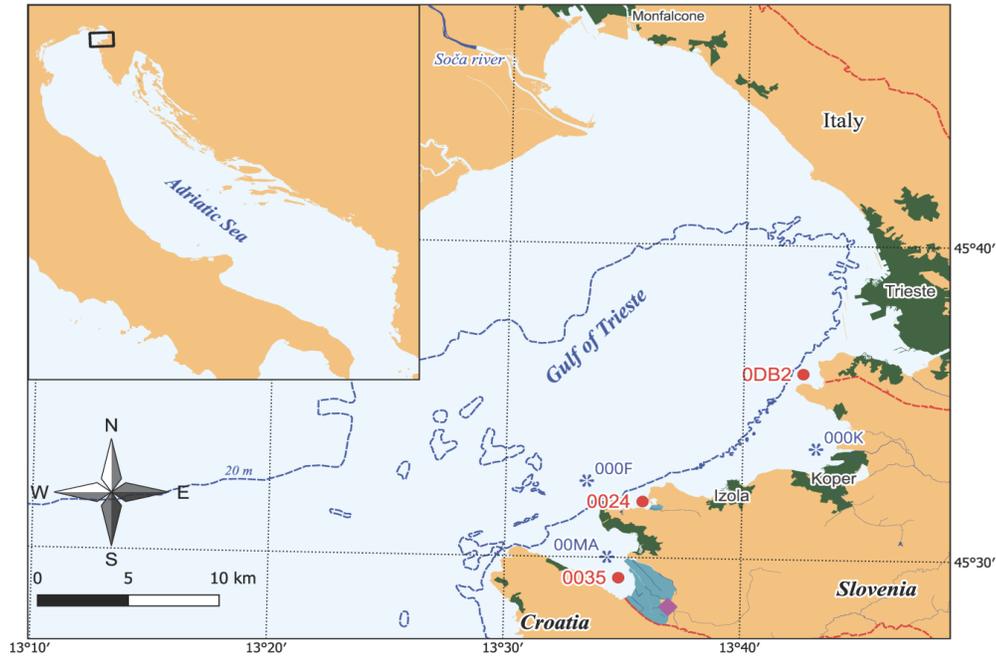

*Figure 1: Map of the GoT (Adriatic Sea) and location of the three mussel farming sites with the respective sampling stations on the Slovenian coast: Debeli rtič (0DB2), Strunjan (0024) and Seča (0035). Physical and chemical parameters of seawater were collected at sampling stations 000K, 000F and 00MA as part of ecological monitoring in accordance with the Water Framework Directive 2000/60/EC. The location of the Portorož Airport meteorological station is marked with a diamond.*

### 2.2. Data acquisition

The data, covering the period 1994-2021, were categorised into three groups according to the type of data and the stations where they were measured.

*Toxic phytoplankton and DSP toxins.* The seawater samples for the phytoplankton analysis and the mussels were collected at three sampling stations, 0024, 0035 and 0DB2, each located in a mussel farming area (Figure 1). For phytoplankton, the sampling period within the year, frequency and sampling method (Niskin bottles at discrete depths, vertical haul with plankton net) varied until 2008, while the program followed a consistent scheme thereafter (complete annual survey, weekly to monthly frequency depending on the season, integrated water sample with a PVC hose). These differences were taken into account when preparing the data for the ML models. The identification and enumeration (cells l$^{-1}$) of the HAB species was performed under the inverted microscope according to the method of Utermöhl (1958). Since we applied ML models to predict DSP events, which are the main problem in the area, we narrowed the selection of data to the five most abundant and recurrent DSP-producing species (*Dinophysis caudata, D. fortii, D. sacculus, D. tripos, Phalacroma rotundatum*), together with the abundance of all DSP-producing species found in seawater samples (*Dinophysis* spp., *Phalacroma* spp. and *Prorocentrum lima*).

A similar approach was used for toxins in mussels. Initially, the sampling program varied considerably from year to year and was often adapted to the results of the phytoplankton analyses, with less regular and frequent sampling in winter during the first decades of observation. From 2011, the scheme became uniform and, as with phytoplankton, covered the whole year, with more frequent sampling in the months of higher toxicity risk (Apr-Nov). Until 2014, the standard mouse bioassay method (AOAC Official Method 959.08) was used in accordance with Directive 91/492/CEE (Council of the European Communities 1991) for the determination of DSP toxins, i.e. okadaic acid and its derivatives, the dinophysistoxins, which are extracted together with other lipophilic toxins such as yessotoxins, pectenotoxins and azaspiracids. The mouse bioassay, having various disadvantages, including ethical concerns, and low sensitivity and specificity with only two possible test results - positive or negative, was in 2014 replaced by the more sensitive and specific method of liquid chromatography-mass spectrometry (LC-MS/MS). The LC-MS/MS method allows chromatographic separation of the individual groups of lipophilic toxins. In Slovenia, harvesting of bivalve molluscs is banned if the regulatory limit of 176 µg kg$^{-1}$ for DSP toxins (160 µg OA equivalents kg$^{-1}$ according to Regulation (EC) No 853/2004 (European Council 2004) plus measurement uncertainty of the national reference laboratory) is exceeded.

*Physical and chemical properties of seawater.* In addition to the two datasets described above, seawater temperature and salinity were also taken into account as they describe the site-specific environmental conditions. Monthly sea surface temperature (SST) and salinity data were obtained from the same sampling stations as for toxic phytoplankton (0DB2, 0024, 0035) or, in case of missing data, from the nearest sampling stations (marked with an asterisk in Figure 1) included in the ecological monitoring according to the Water Framework Directive 2000/60/EC. Temperature and salinity were measured with a fine CTD probe (Sea & Sun Technology GmbH).

*Meteorological and hydrological data.* Meteorological observations such as daily average air temperature and wind speed, daily solar irradiance and daily precipitation at the coastal meteorological station (Porotož Airport) were obtained from the Slovenian Environment Agency[‡], which also provided the daily averaged flow rates of the main freshwater source, the Soča River.

### 2.3. Data preprocessing

The performance of an ML model depends largely on the quantity and quality of the data used to train it. The right choice of input variables affects the predictive performance of the model, as does an imbalance between classes (Menardi and Torelli 2014), which can lead to biased learning because the model learns to predict the majority class better. Our goal was to collect all relevant input variables from the different data sources (e.g. Cruz et al. 2021; Patrício et al. 2022), while obtaining a sufficient amount of data for ML training after preprocessing the missing values. We applied extensive data matching, aggregation and interpolation methods to obtain a consolidated dataset with 14 independent variables and DSP toxicity as the target variable, which were then used for training (Table 1).

---

[‡] www.arso.gov.si

| Variable | Description | unit | Min | Max | Mean | Median |
|---|---|---|---|---|---|---|
| month | | | | | | |
| DSP-tot | | cells l$^{-1}$ | 0 | 7634 | 96 | 30 |
| *Dinophysis caudata* | | cells l$^{-1}$ | 0 | 1309 | 23 | 0 |
| *Dinophysis fortii* | | cells l$^{-1}$ | 0 | 4624 | 23 | 0 |
| *Dinophysis sacculus* | | cells l$^{-1}$ | 0 | 4639 | 26 | 0 |
| *Dinophysis tripos* | | cells l$^{-1}$ | 0 | 1139 | 4 | 0 |
| *Phalacroma rotundatum* | | cells l$^{-1}$ | 0 | 393 | 13 | 5 |
| SST | | °C | 6.23 | 28.87 | 17.41 | 17.28 |
| salinity | | | 24.13 | 38.66 | 36.54 | 36.99 |
| air temperature | average air temperature in 20 days prior to phytoplankton sampling | °C | -1 | 27 | 17 | 18 |
| wind speed | average wind speed in 20 days prior to phytoplankton sampling | m s$^{-1}$ | 1 | 5 | 3 | 3 |
| precipitation | daily precipitation summed over 20 days prior to phytoplankton sampling | mm | 0 | 268 | 59 | 45 |
| solar irradiance | daily solar irradiance summed over 20 days prior to phytoplankton sampling | h | 27 | 278 | 164 | 168 |
| river flow | daily flow rate of the Soča River summed over 30 days prior to phytoplankton sampling | m$^3$ s$^{-1}$ | 594 | 16040 | 3301 | 2580 |
| DSP toxins | positive or negative result of toxicity test | | | | | |

*Table 1: Input variables for the training set with descriptive statistics over 28 years (1994 - 2021).*

The abundances of five phytoplankton species and the aggregated variable DSP-tot – the sum of the abundances of all DSP-producing species – were included. As sampling techniques changed during the monitoring program, the abundances from the net samples were increased by two orders of magnitude for better comparability. When multiple samples were from the same day and location but from different depths, the samples with the highest abundance of DSP-tot were selected.

The toxicity results also had to be standardised to take account of the different methods. In accordance with the legal threshold for DSP toxins (176 μg kg-1), the concentrations of the chemical analyses (2014-2021) were mapped to binary positive or negative test results to match the results of the bioassays (1994-2013). The total number of toxicity tests received (binary target variable) was 1132, of which 996 (88%) were negative and 136 (12%) were positive.

As the phytoplankton and toxicity datasets had different sampling frequencies and temporal resolutions, an appropriate toxicity test was assigned to a phytoplankton observation based on a given time window, separately for the three sampling stations. A Python script selected the first possible toxin result based on the timestamp of the phytoplankton observation, but only if it was not older than 30 days.

Before we could proceed with the aggregation of SST and salinity, the problem of missing data had to be solved. This problem was mitigated by interpolation, where missing data were replaced by data from measurements at the nearest location (either mussel farming stations or WFD monitoring stations) and within a certain time window (+/- 30 days). This method avoided potentially erroneous approximations and replaced 1215 missing values (38% of missing values). For the meteorological data, daily solar radiation, precipitation, air temperature and wind speed were added by averaging them for a time window of 20 days prior to the phytoplankton observation. Finally, the daily flow of the Soča River was summed over a period of 30 days prior to the phytoplankton observation.

The dates in the consolidated dataset were converted to months to allow the model to better capture the annual cyclical patterns of phytoplankton growth. Through these steps, the consolidated dataset retained all 1452 instances from phytoplankton monitoring and, importantly, all toxicity results (1132). In a further preprocessing step, instances with missing values (44 in total) were removed, as some ML algorithms cannot process data with missing values.

We used UMAP (McInnes, Healy, and Melville 2018) for dimensionality reduction and visualisation in two dimensions to gain insight into how the data are partitioned with respect to the target variable. This method projected the consolidated dataset into a lower-dimensional feature space, attempting to preserve as much variance or structure as possible. The positive and negative examples were quite mixed throughout the data space (Figure 2, left), making it more difficult for a model to learn a clear decision boundary between the data. Since the dataset was heavily imbalanced in favour of the negative class, we used the Edited Nearest Neighbours from the imbalanced-learn Python toolbox (Lemaître, Nogueira, and Aridas 2017) to remove instances from this class whenever they were close to instances with positive class values. This removed 211 "conflicting" examples with negative test results (Figure 2, right).

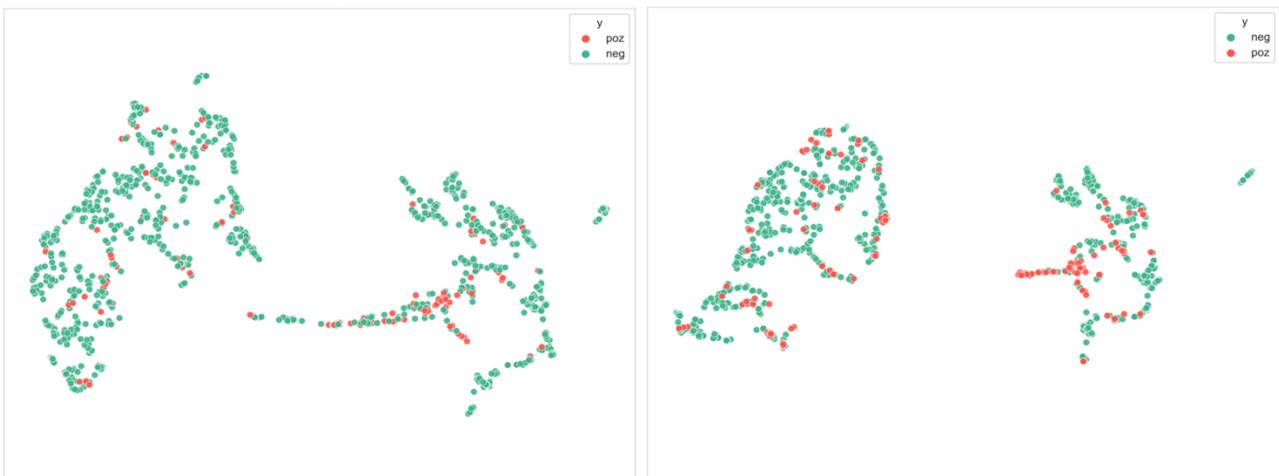

*Figure 2: UMAP projection of data in two dimensions before (left) and after (right) removing instances that are closely positioned in data space and have opposite class values.*

The resulting dataset comprised a total of 877 instances (target class distribution: 745 negative, 132 positive). The dataset was then randomly split into a training and a test dataset, with a 70/30 split and stratified sampling to maintain the proportion of negative/positive examples in both datasets.

### 2.4. Training machine learning models

ML algorithms were used to train models for the prediction of DSP events in two binary classes - positive (1) or negative (0). One factor in the selection of ML algorithms and in determining the hyperparameters was the relatively small number of instances, especially of examples with positive DSP. In addition, the goal was to obtain ML models that could be directly explained and to use explainability methods for more opaque models. This of algorithms to tree-based algorithms such as decision tree (DT) and random forest (RF), support vector machines (SVM) and shallow artificial neural networks (ANN). These ML algorithms are widely used for ecological modelling and also for the prediction of HABs and DSP events (Grasso et al. 2019; Liu et al. 2022; Kim et al. 2021).

DT is particularly useful in scenarios where interpretability is crucial and the problem is not overly complex as it builds decision rules that can be easily understood by aquaculture operators and other stakeholders. RF is widely used in ecological modelling because it can capture more complex nonlinear relationships and provides some explainability with built-in methods (feature ranking). SVM is particularly effective in dealing with high-dimensional data and complex relationships between variables when the data have a clear margin of separation. In the last decade, ANNs have become popular, especially deep learning methods, as they can model very complex nonlinear phenomena with potentially high predictive performance. However, they require a large amount of training data, are prone to overfitting and are considered black boxes as the decisions of the models are opaque (Recknagel 1997). Due to the limited amount of training data, a shallow feedforward neural network, the multilayer perceptron (MLP), was used. This algorithm required further preprocessing of the data as it works better when the data is standardised, for which z-score normalisation was used. The month variable was also removed.

As the training set was significantly imbalanced, we considered addressing this issue by undersampling and data augmentation. As Kim et al. (2021) have shown, synthetic data can improve the ML models' detection of HABs. Our method integrates under- and oversampling into the modelling pipeline, which systematically explores and optimises hyperparameters with a grid search over a predefined range of values to determine the most effective settings for data augmentation and ML algorithms. Criteria for splits, class weight adjustments and structural parameters such as tree depth and layer complexity were taken into account. This careful process ensured robustness against overfitting, especially given the constraints imposed by the size of the dataset.

In the first step, the training data (validation data) was oversampled using the Synthetic Minority Over-sampling Technique (SMOTE) to increase the number of instances with positive toxicity. The next step in the pipeline was to undersample the majority class (negative toxicity) of the training data to further reduce the class imbalance. In the third step, the ML models were trained with stratified 5-fold cross-validation on the over- and undersampled training set with the optimised parameter values. While several model performance metrics were calculated (recall, precision, F1 score), the optimal algorithm parameter configuration was determined based on the mean F1 score of each grid search combination over the 5 folds.

## 2.5. Evaluation of models

To assess the generalizability, the evaluation of the constructed ML models was performed on unseen examples and on data distributions they would encounter in real-world applications. Based on the performance of the obtained models, the highest-ranking model from grid search was then selected for each of the ML algorithms and subsequently evaluated on the test set that has been left out of the training pipeline. This step was repeated 100 times, and the results of the selected metrics (precision, recall, F1-score) were averaged to obtain a more statistically reliable estimate of each model's performance. The most relevant metric for the task at hand was recall, as it indicates the proportion of predicted positive class instances relative to all true positive (TP) instances in the test set. However, to ensure that the classifier does not predict every instance as positive and thus make too many false positive (FP) predictions, we also needed to control precision. Therefore, the F1 score was used in the parameter optimisation as it balances both recall and precision. Finally, the averaged performances for all three performance metrics were compared for all four ML algorithms.

## 2.6. Explaining ML models

This study places the same emphasis on the predictive power as on the explainability of the model. By applying XAI principles, the aim is to better understand the inner workings of models by identifying the most informative relationships between the input variables and the predicted target variable. XAI calls for the use of interpretable ML models whenever possible and the application of explainability techniques for opaque ML models: Shapley Additive Explanations (SHAP) (Lundberg and Lee 2017), Local Interpretable Model-Agnostic Explanations (Lime) (Ribeiro, Singh, and Guestrin 2016), permutation importance, various feature ranking methods, and others.

The DT used in the study could be inspected directly by visualising the tree structure of the model. For the more complex models – RF, SVM and ANN – selected explainability methods were used to gain insights into their behaviour. The importance of the variables, i.e. the feature importance, was determined using two selected model-agnostic methods that can be applied to different model types. Permutation feature importance from the scikit-learn library was used to obtain a ranking of how strongly each of the variables influences the RF model. This permutation-based method of variable importance reflects the decrease in a model's performance score when a single variable value is randomly shuffled (Breiman 2001). Shuffling removes the relationship between the independent and the target variable, leading to a decrease in model performance and thus showing how much the model depends on the particular variable.

The SHAP method was used to gain a better understanding of the contribution of the individual variables to the models' output. For RF, the SHAP TreeExplainer, an algorithm specifically for tree ensemble methods, was applied. This method allows a general model interpretation as it represents the behaviour of the model over the entire data set on which it was trained. In addition, it also allows us to inspect individual model's predictions which were also implemented for this study, as they are relevant for real-world deployment.

### 2.7. Implementation

The data preprocessing, the statistical analysis, the parameter optimisation and the model construction, evaluation and interpretation were carried out using the Python programming language. In the study, it was used used in combination with JupyterLab, an interactive development environment for computational notebooks (i.e. Jupyter notebooks), and pandas (McKinney and Others 2010). ML algorithm implementations from scikit-learn (Pedregosa et al. 2011)and other specialised tools such as UMAP (McInnes, Healy, and Melville 2018), Imbalanced-learn (Lemaître, Nogueira, and Aridas 2017) and SHAP (Lundberg and Lee 2017) were used. The complete code and data are available in an online repository[§].

## 3. Results

### 3.1. Analysis of long-term phytoplankton and DSP toxins monitoring

The basic oceanographic features in the Slovenian part of the GoT display notable annual fluctuations in seawater temperature and salinity, as shown in Figure 3. From 1994 to 2021, SST fluctuated between 6.23°C in February and 28.87°C in July, while surface salinity fluctuated between 24.13 and 38.66 (Table 1), with the two extremes occurring in February. The lowest monthly salinities (shown as triangles in Figure 3) were measured in the winter months (January-February) and in November, suggesting that less saline water is present in the surface layer during these periods. However, on average, the winter months had high salinity, while the late spring/summer months (May-July) and November had below average salinity. The highest temperature peaks for each month (also indicated by triangles in Figure 3) were mostly observed in the last decade. The largest deviations from the average climatology were observed in spring/summer (May-July).

---

[§] GitHub: https://github.com/MartinMarzi/HABTox-predictor

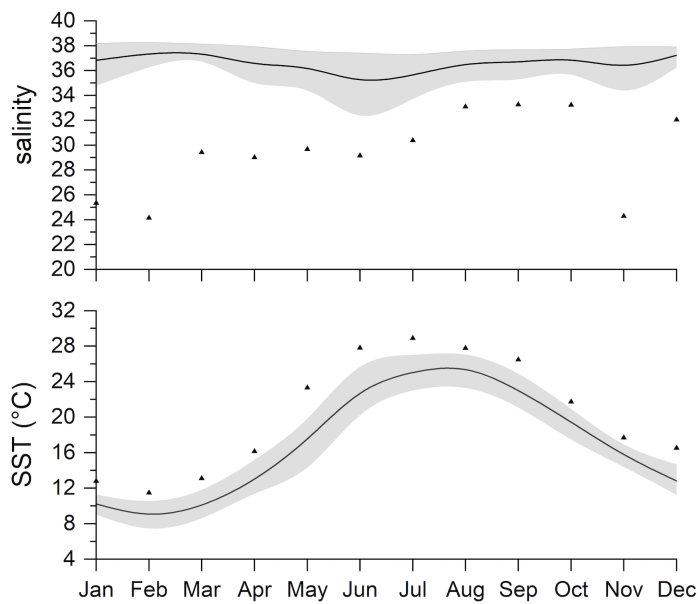

*Figure 3: Annual distribution of mean salinity and seawater temperature (SST) in the surface layer of Slovenian coastal waters, period 1994-2021. The grey band around the mean (line) is the 10-90 percentile. The triangles indicate the lowest salinity and the highest temperature for a given month measured at any point during the 28-year time series.*

The annual distribution of DSP-tot (as 90th percentile), averaged over the 28-year time series, shows that the period from May to December is the most likely time for DSP outbreaks in the GoT (Figure 4). Two peaks in abundance, reaching values of up to 7600 cells $l^{-1}$ (Table 1), were observed in June/July and from September to November, corresponding to the distribution patterns of the five main species. The "blooms" of *Dinophysis sacculus* and *D. caudata* were typical of the early summer months, while *D. fortii* was responsible for the fall peak. During the rest of the year, the DSP-producing species were almost absent in seawater.

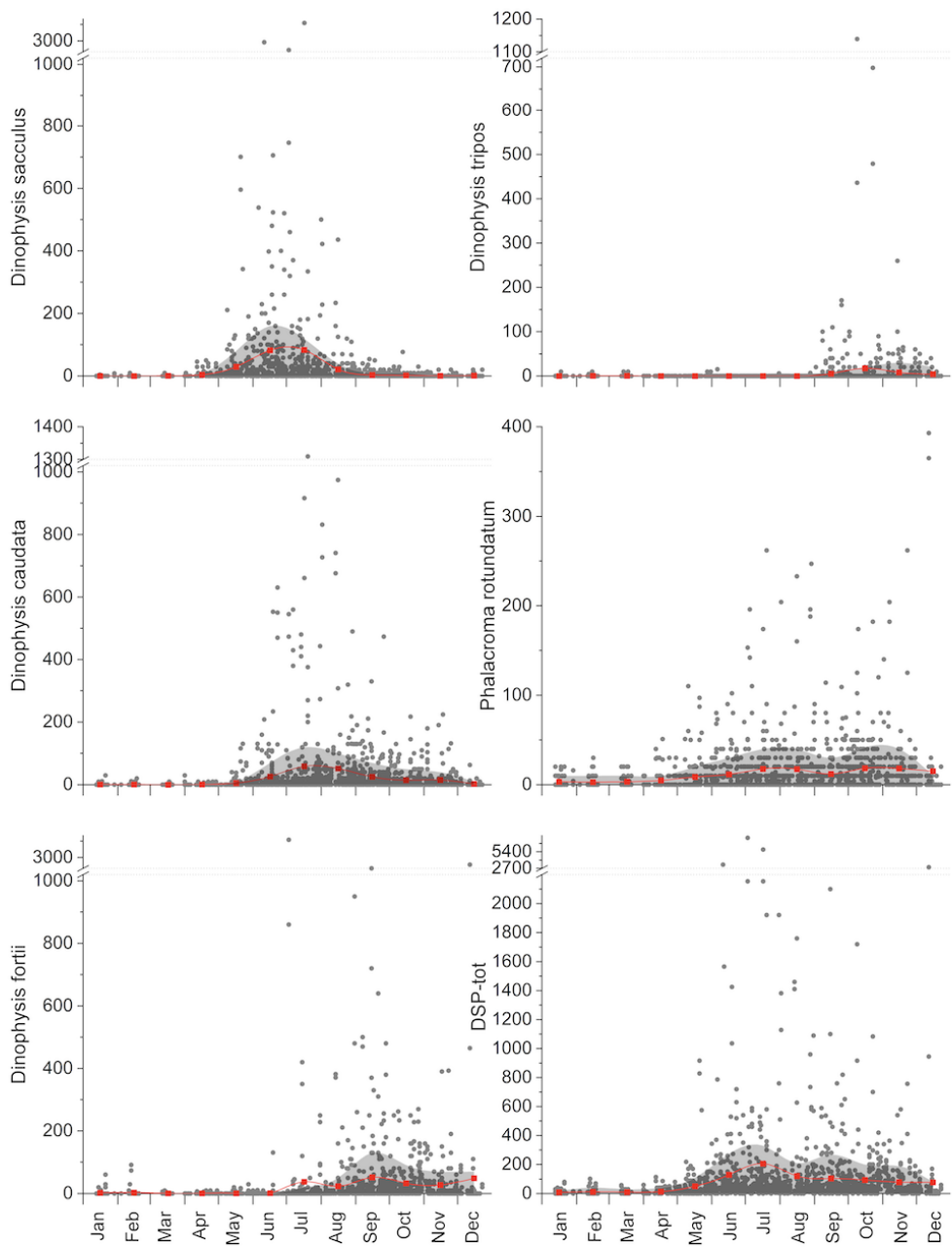

*Figure 4: Annual distribution of abundance (cells l$^{-1}$) of the five DSP-producing species and DSP-tot used as independent variables in ML models. The dots represent individual observations, the grey area the 90th percentile, while the red line represents the mean for the period 1994-2021. Note the different scaling of the y-axes.*

The DSP toxin analyses were intensified from May to November to follow the dynamics of the toxic species. Monthly testing varied greatly, ranging from about 20 tests in winter to about 170 tests in fall from 1994 to 2021 (Figure 5). The number of positive tests increased significantly in the months with increased sampling, especially in September and October, where absolute values peaked at 31 and 24 tests, accounting for 23 % and 18 % of all positive tests, respectively. The incidence of DSP poisoning was lowest from December to April (2% of all positive tests).

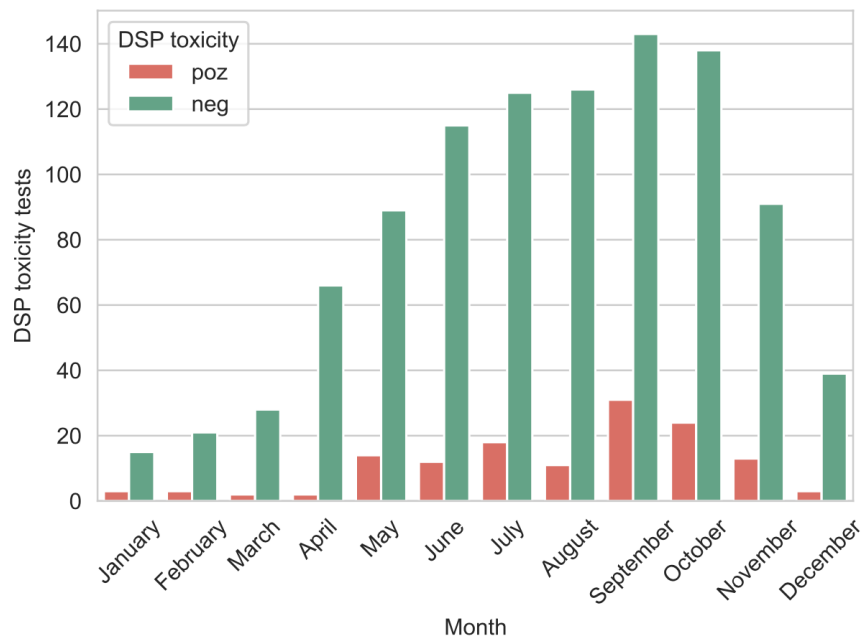

*Figure 5: Annual distribution of DSP toxin test results by month, over the years 1994-2021.*

### 3.2. Training ML models and performance evaluation

Using the ML pipeline, SVM, DT, RF and ANN models were trained to predict DSP toxicity, with hyperparameters individually optimised using grid search based on the mean F1 score over 5 folds. Table 2 gives an overview of the best hyperparameters for each ML algorithm.

|  | SVM | DT | RF | ANN |
|---|---|---|---|---|
| **model** | C = 100<br>Class_weight = None | Class_weight = None<br>criterion=entropy<br>max_depth=4 | Class_weight=None<br>criterion=gini<br>N_estimators = 300 | activation=relu<br>max_iter=5000<br>hidden_layer_sizes = 3<br>batch_size=min(200, n_samples) |
| **SMOTE** | k_neighbors=3<br>sampling_strategy=0.4 | k_neighbors=3<br>sampling_strategy=0.3 | k_neighbors=3<br>sampling_strategy = 0.4 | k_neighbors = 5<br>sampling_strategy = 0.6 |
| **RandomUnderSampler** | sampling_strategy=0.6 | sampling_strategy=0.6 | sampling_strategy=0.5 | sampling_strategy = 0.7 |

*Table 2: The best hyperparameters of the ML algorithms determined using grid search.*

To obtain a reliable estimate of how the models would perform with similar data in the real world, the entire pipeline (model construction with hyperparameter optimization on the training set and evaluation on the test set) was run through 100 iterations and the results averaged. This was important due to the variability of results between runs and between folds (Figure 6) and allowed for a more reliable algorithm comparison using the three performance metrics of precision, recall, and F1 score. Table 3 shows the overall performance of the models.

| Classifier | Precision | Recall | F1-score |
|---|---|---|---|
| SVM | 0.42 | 0.47 | 0.43 |
| DT | 0.42 | 0.48 | 0.43 |
| RF | 0.74 | 0.59 | 0.65 |
| ANN | 0.58 | 0.45 | 0.49 |

*Table 3: Average model test results for four ML algorithm classes over 100 iterations.*

The results show that the ML models are able to predict DSP toxicity. RF had the highest and most stable prediction performance over the 100 iterations with an average precision of 0.74 (std ±0.09), recall of 0.59 (std ±0.08) and F1 score of 0.65 (std ±0.07). The study found that among the selected ML algorithms for this particular type of data, RF is the most suitable model for the direct prediction of DSP toxicity in mussels. Interestingly, the less complex DT performed on average very similarly to the ANN, with both having an identical precision of 0.42 and F1 score of 0.43, while the ANN had a slightly higher recall of 0.48 (DT 0.47).

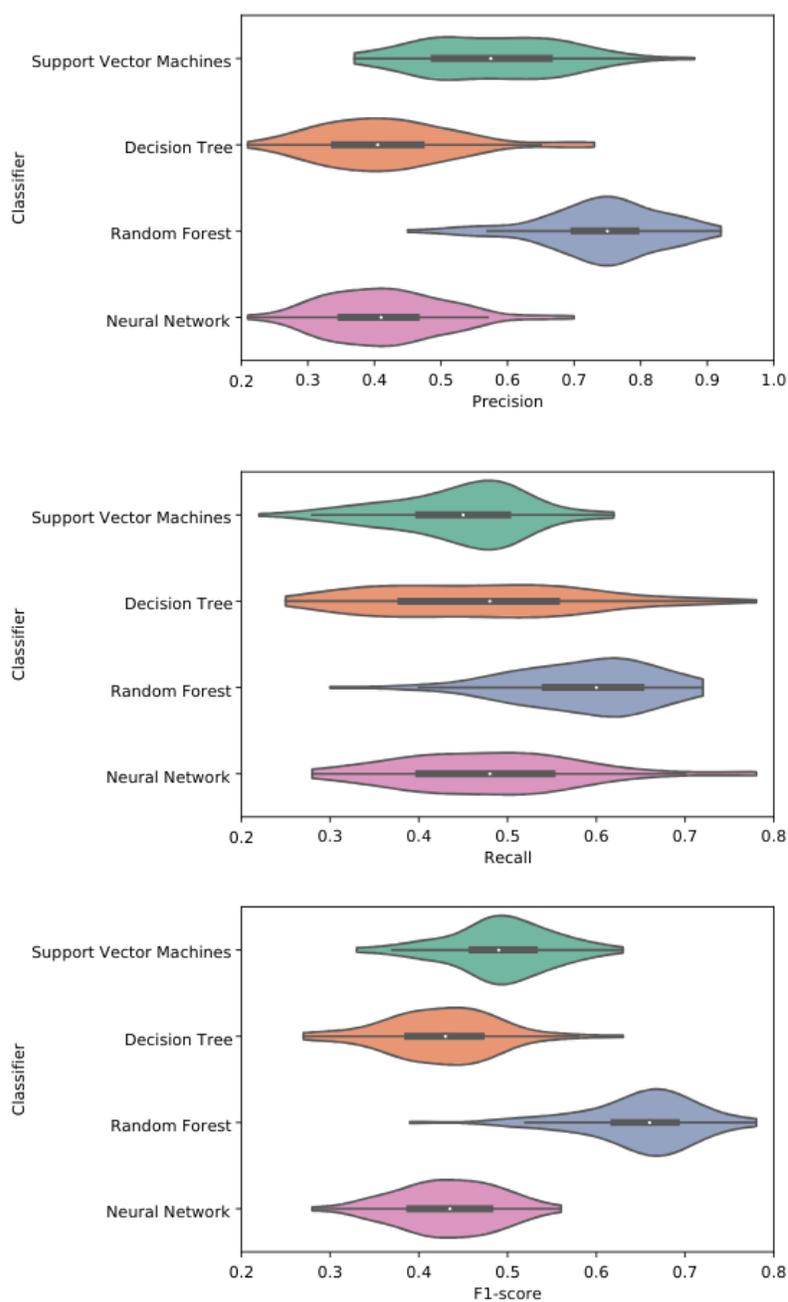

*Figure 6: Variability of performance metrics over 100 iterations.*

It is important to note that the models were optimised for the F1 score during parameter optimisation and that models were constructed with significantly higher recall (Figure 6), but at the expense of lower precision. For the purposes of the study, a balanced approach to recall and precision was taken when evaluating the performance of the models. While the highest possible recall that would predict the most positive toxicity examples, would seem favourable for use in real EWS, it would also produce too many FP. This is also evident in the precision-recall curve for the RF model in Figure 7. As can be seen, the precision drops steeply at higher recall values above approx. 0.6.

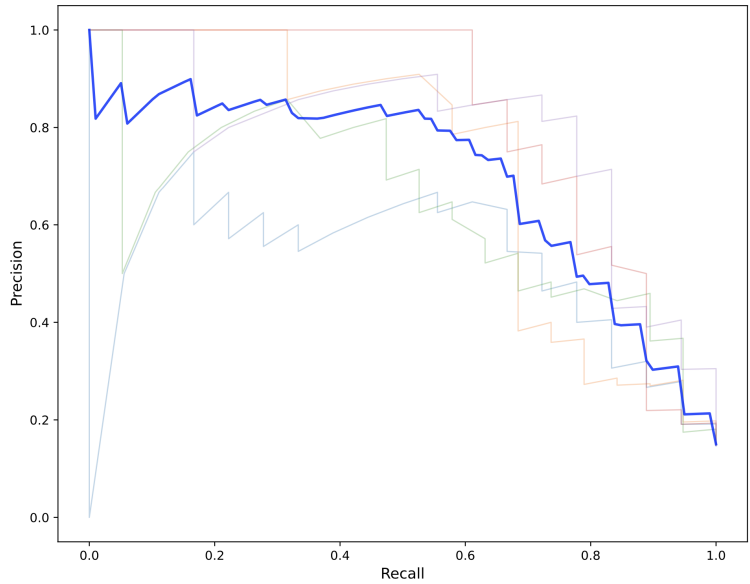

*Figure 7: Precision-recall curves for the best RF model of grid search (blue curve is the average). At higher recall values (>0.6), the precision drops steeply.*

### 3.3. Explaining machine learning models

An example of the constructed DT with optimised parameters from grid search (Table 2) and trained on all data (training and test data combined in order to take advantage of all available data) is presented in Figure 8. In this example, the decisions of the model can be directly interpreted. This gives a good insight into how the model makes predictions. At the first step the model splits the data according to the presence of *D. fortii* at abundances above 30 cells $l^{-1}$. If the abundance is higher the model predicts positive test results. With abundances equal or lower than 30 cells $l^{-1}$ the decision depends on the presence or absence of *D. caudata*. When *D. caudata* is present, the next decisive variable is salinity, which leads to positive toxicity results at values ≤ 36.17 while at higher salinity the prediction is always a negative toxicity result.

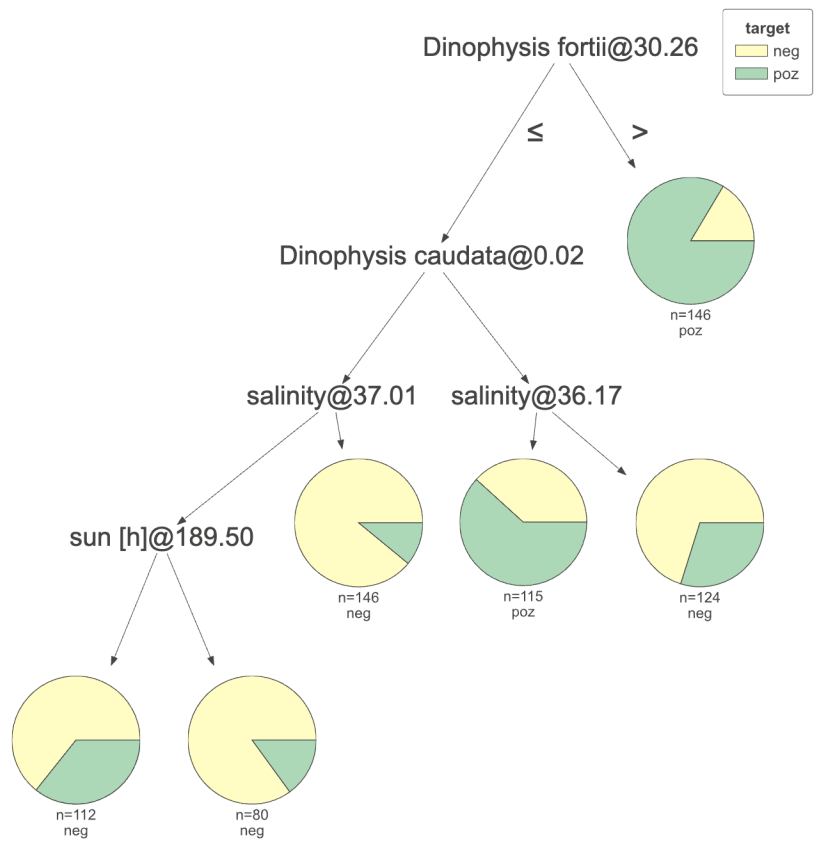

*Figure 8: Visualisation of a constructed DT model to illustrate the decision rules for DSP toxicity results. The numbers at each leaf node (pie chart) indicate the total number of remaining instances. The prediction of the model is labelled "poz" for positive toxicity predictions and "neg" for negative toxicity predictions.*

To investigate the behaviour of a typical RF model that performed best among the opaque models (F1 score), we applied two explainability methods. Figure 9 shows the permutation importance, with the variables ordered from top to bottom. The length of each bar indicates how much the model performance decreases when the values of the respective variables are randomly shuffled. In the RF model presented, *D. fortii* is the variable with the greatest influence, followed by DSP-tot and Soca river flow.

Variables in the lower part of the diagram with negative values indicate that the predictions for the shuffled data were more accurate than the original data. However, with small data sets, as is the case here, such a result may occur more frequently due to chance. Therefore, these variables have little or no significance for the predictions of the RF model according to the permutation importance method.

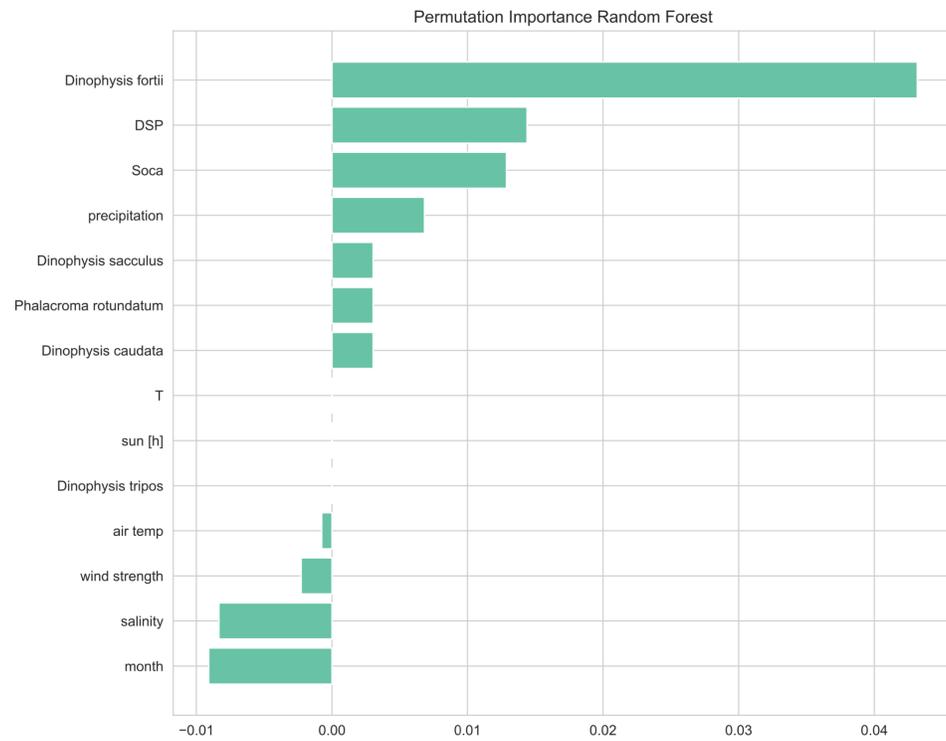

*Figure 9: Permutation importance of a typical RF model showing the ranking of the relevance of the independent variables for the prediction of DSP toxicity.*

Next, SHAP was used to explain the results of the RF model. In Figure 10, the beeswarm plot summarises the overall distribution of Shapley values for each variable and example (point) in the test set. A positive Shapley value indicates that the presence of the variable increases the probability of the target class compared to the average prediction. Conversely, a negative Shapley value means that the feature lowers the prediction value. Larger absolute Shapley values indicate that the feature has a stronger effect on the prediction, while a Shapley value close to zero means that the feature has little to no effect. The beeswarm plot provides an overview of which variables are most important for the predictions of the RF model by ordering the variables according to the sum of the absolute Shapley values for each variable in the test set. The distribution of the variable values (colour scale in Figure 10) along the axis of the Shapley value is important for interpreting the importance of the variables. *D. fortii* is ranked first on the beeswarm plot, which, together with the distribution of red dots on the positive side, indicates that higher abundance of this species influences the prediction of the RF model by increasing confidence in a positive toxicity result. Similar observations can be made for the next two most important variables, *D. caudata* and DSP-tot. In contrast, lower flow rates of the Soca river and higher salinity levels reduce RF confidence in a positive test result. A lower air temperature also appears to influence the model in such a way that the prediction probability for a positive toxicity result decreases. Other variables with Shapley values closer to zero have less influence on the model's decision.

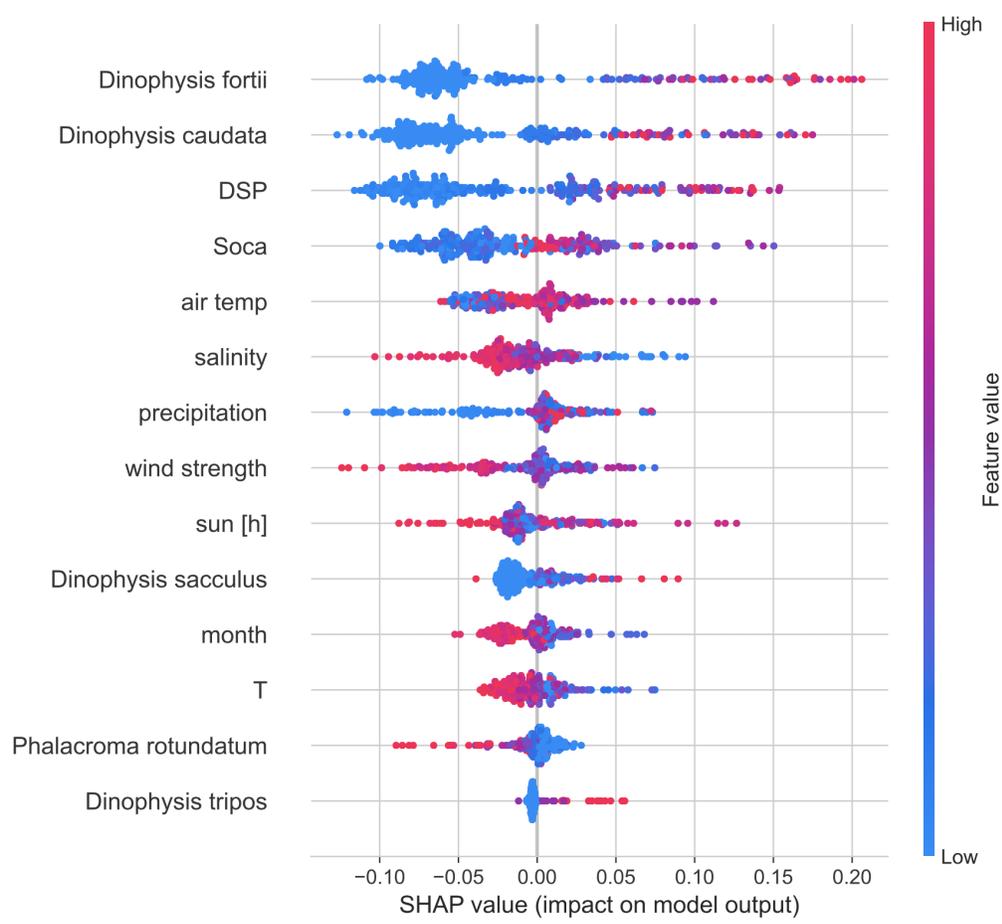

*Figure 10: The summary beeswarm plot of the Shapley values for the independent variables of each instance in the test set predicted by a typical RF model.*

SHAP was also used to explain the model's prediction for a selected example. In Figure 11, the SHAP force plot illustrates how much each independent variable is contributing to push the output from the base value (0.33) to the actual output produced by the RF model. Variables forcing the prediction higher are shown in red, while those forcing the prediction lower are in blue. As can be seen, RF correctly predicted this example as positive (predicted value = 1) with a confidence of 0.58.

The high abundance of DSP-tot (460 cells $l^{-1}$), to which *D. fortii* (330 cells $l^{-1}$) and *D. tripos* (110 cells $l^{-1}$) contributed the most, and the long duration of solar radiation (187.5 h) have the highest relevance for the prediction result, forcing it to have higher confidence values. In contrast, the low flow rate of the Soča river (1174 m3 $s^{-1}$) and the high salinity (37.67) are the main variables that reduce the model's confidence in a positive prediction.

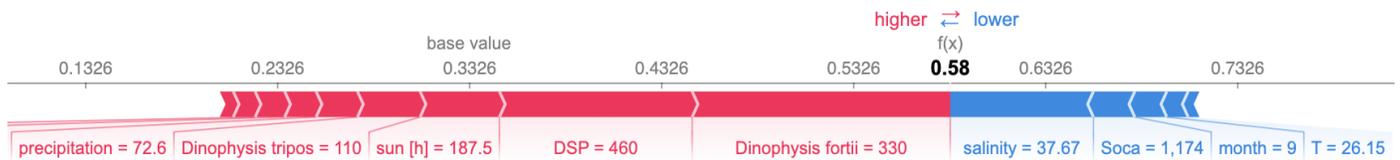

*Figure 11: SHAP force plot of individual RF model prediction of DSP toxicity (names of some less impactful variables on the left – air temp,* D. caudata *and* Phalacroma rotundatum *– are not shown).*

## 4. Discussion

The complexity and non-linearity of the ecological dynamics responsible for DSP events results from a multitude of biological and environmental factors, as well as from the not well understood contamination/decontamination kinetics of mussels (García-Corona et al. 2022). Such an interplay of factors requires advanced predictive models (Cruz et al. 2021) and the right selection of variables (Yu et al. 2021). In this study, several ML algorithms were used for modelling and the five most widely distributed DSP species in Adriatic coastal waters which pose a significant risk of DSP toxin contamination in mussels, were included (Henigman et al. 2024; (Ninčević Gladan et al. 2020)). Dinophysis species are known to thrive in a stable water column stratification, which is often associated with increased seawater temperature and reduced surface salinity due to precipitation or freshwater discharges. In addition, local hydrodynamic features such as upwelling and downwelling cycles, tides and coastal advection influence Dinophysis populations (Reguera et al. 2012). Therefore, variables such as seawater temperature, salinity, turbidity, chlorophyll-a, wind, air temperature and precipitation appeared to be relevant, and they have also been used in other studies to predict toxicity with ML, including DSP (Capoccioni et al. 2023), PSP (Harley et al. 2020), or both (Bouquet et al. 2022). Although not investigated in our study, further physiological aspects of phytoplankton such as toxin production dependent on abiotic factors (e.g. temperature) or intrinsic factors (growth phase) together with bivalve physiology (e.g. bioaccumulation processes) could improve model performance.

The constructed dataset presented several challenges for training ML models. There was a significant class imbalance in the DSP toxicity variable, as events with high toxicity values occurred disproportionately less frequently than events with low values. Consequently, data augmentation techniques were incorporated into the model training pipeline to mitigate this. In addition, dimensionality reduction revealed that instances with positive and negative toxicity values were spread across the dataspace (Figure 2), suggesting that some of the data sources were noisy (possible reasons include measurement and processing errors as well as changes in monitoring methods during the long time span). Since this causes the ML models to have difficulty in drawing a decision boundary, we removed instances belonging to the negative class whenever they did not match with nearby instances belonging to the positive class. In this way, the most "conflicting" instances of the majority class were removed and the target classes were further balanced. Although this did not completely eliminate the underlying problem, this approach had a positive effect on the performance of the model.

Despite the challenges, model evaluation demonstrated the effectiveness of ML in predicting DSP toxicity in mussels, with RF algorithms outperforming other models due to their robustness and ability to handle complex interactions within the data (Table 3), which is consistent with findings on the effectiveness of ensemble methods on similar problems (Cruz et al. 2021; Harley et al. 2020). Ensemble methods are known for their variance reduction and resilience to noise and outliers. In the evaluation, RF showed an average recall of 0.59, with most models exceeding this value (Figure 6) despite being optimised on the F1 score. When parameters were optimised, higher recall rates were observed, but for a balanced evaluation, models with better F1 scores were preferred. When implementing EWS, prioritising recall could improve the detection of positive toxicity at the expense of higher FP — a trade-off considered acceptable due to the likely subsequent verification of toxicity in the laboratory.

However, the evaluation of the robustness of the model was limited by the small size of the test set, which urges caution when applying these models to different data sources. ML models may have limited generalizability and may perform suboptimally on data that deviate from the training set. In general, it should be kept in mind that site- and species-specific models are superior to generalizable models (Rousso et al. 2020).

By applying XAI approaches in marine ecology, our study aimed to fill important gaps in ML applications in this field while emphasising the need and potential of XAI. Especially for scientific and real-world applications, – such as EWS–, it is insufficient to only estimate how reliable the predictions of a model are. For an ML-based EWS to be perceived as trustworthy, authorities also need to understand the rationale behind the model's decisions so that they can make an informed decision on appropriate actions.

Our study shows how XAI techniques can provide insights into the model's decisions by revealing the correlations learned by the ML algorithms. In two different explainability methods (Figures 9 and 10) of the best performing ML model (RF), *D. fortii* was ranked as the most influential variable for positive toxicity prediction, followed by the entire DSP assemblage or specific species such as *D. caudata*. These results are largely consistent with real data from in vitro studies demonstrating the synthesis of DSP toxins in isolated cells of *D. fortii* (Yasumoto et al. 1980; M. G. Park et al. 2006) or from field studies. Our data show that the period with the highest abundance of *D. fortii* coincides with the periods with the highest incidence of DSP above the regulatory limit in September and October (Figures 4 and 5), while earlier events of DSP outbreaks, from May to August, could be due to *D. caudata* or *D. sacculus*, which also increased the probability of positive test results. In addition, a recent study from Slovenian coastal waters, conducted with a much shorter dataset (6 years), showed the strongest association between DSP toxins in mussels and *D. fortii* (Henigman et al. 2024).

The next highly ranked variables are those related to the supply of freshwater to the coastal waters, such as the flow rate of the Soča River and precipitation. These variables alter the hydrodynamics and increase the stratification of the water column by reducing salinity in addition to the high water temperatures during warm periods. Such conditions can promote the growth of Dinophysis species and their accumulation in thin layers. The interplay between these influential abiotic factors and the dynamics of DSP species, particularly *D. fortii*, is also evidenced by a real case of mussel toxicity that was correctly predicted by the model with relatively high confidence (Figure 11). The factors that increased the predictive confidence of the model the most were the high abundance of *D. fortii* and DSP-tot (330 cells l$^{-1}$ and 460 cells l$^{-1}$ respectively are among the highest values in our study area) and the long day length, serving as a proxy for warm, stable summer conditions. Conversely, high salinity, low cumulative river flow and the month of September had the opposite effect and reduced confidence in the model's positive prediction. While the first two factors were correctly identified as least influential for a positive prediction, the influence of the month led to a bias in the model result, as the highest number of toxic events occurred in September (see Figure 5). Although the drivers mentioned above clearly influenced the predictions of the model, this does not prove causality as correlation does not imply causation.

Although the DT model exhibits lower performance than the RF model, it also identifies a similar set of variables as RF. However, DT offers better interpretability for potential end users due to its simplicity and ease of use. Our example shows (Figure 8) that even low abundances of *D. fortii* (30 cells l$^{-1}$ or more) can serve as warning indicators of positive toxicity. In cases where *D. fortii* has less than 30 cells l$^{-1}$, the presence of *D. caudata* and salinity levels ≤36.17 should be considered as warning indicators. Despite its suboptimal performance, DT provides thresholds for decisions that can be easily verified retrospectively. When applied to the data of a mussel farm, the informative value was significantly improved from 2013, when the more precise analytical method for the detection of DSP toxins was introduced into the Slovenian national monitoring of mussels.

## 5. Conclusions

The study represents an advance on the less explored and more difficult problem of applying ML to directly predict DSP toxicity in mussels in the Adriatic Sea. To this end, a new dataset of toxic phytoplankton and DSP toxins from the GoT, spanning almost three decades, was created and is openly available for further research. Another focus of the study is the application of XAI principles by using interpretable ML models and explainability methods for opaque models to gain helpful insights into the complex interactions between marine organisms and their environment.

The following main conclusions can be drawn from the study:

- The right data preprocessing steps are crucial for overcoming the specific challenges of consolidated datasets from different sources, and determine model training and performance.
- ML models, especially RF, can satisfactorily predict DSP toxicity in mussels from Slovenian mussel farms.
- Both the RF explainability methods and the DT visualisations show that *Dinophysis fortii* and *Dinophysis caudata* together with the abiotic factors influencing the salinity of coastal waters (river discharge and precipitation) have the greatest influence on predictions.

- The insights about the model's behaviour gained from explainability methods make ML approaches suitable for the EWS due to the increased trustworthiness.
- Predictive performance should be calibrated for the needs of EWS by optimising model training on the performance metric that is most important in the real world (e.g., recall).

The machine learning models developed, when integrated into EWS, can provide a cost-effective means of implementing timely and appropriate mitigation actions, such as trade bans, while improving management strategies to minimise health risks and social and economic damage. Future improvements should focus on refining these models by expanding the training dataset, especially with more positive toxicity tests, and by continuously improving the temporal resolution and quality of the training data through improved monitoring methods. For example, the robustness of these models could be significantly improved by including a wider range of data of high-toxicity events and incorporating datasets from neighbouring regions.

**Declaration of Competing Interest**

The authors declare that they have no known financial interests or personal relationships that could influence the work presented in this study.

**Acknowledgments**

This work was financially supported by the National monitoring program for toxic phytoplankton and marine biotoxins in shellfish growing areas in the Slovenian Sea (The Administration of the Republic of Slovenia for Food Safety, Veterinary Sector and Plant Protection) and by the Slovenian Research and Innovation Agency (grant numbers P1-0237, P4-0092 and P2-0103).